\documentclass[11pt,a4paper]{article}
\PassOptionsToPackage{pdftex,
            pdfauthor={Ikuya Yamada, Hiroyuki Shindo},
            pdftitle={Neural Attentive Bag-of-Entities Model for Text Classification}}{hyperref}
\usepackage[hyperref]{conll-2019}
\usepackage{times}
\usepackage{latexsym}
\usepackage{amsfonts}
\usepackage{amssymb}
\usepackage{graphicx}
\usepackage{multirow}
\usepackage{float}
\usepackage{url}

\aclfinalcopy

\title{Neural Attentive Bag-of-Entities Model for Text Classification}

\author{
    \begin{minipage}{13em}
        \begin{center}
            Ikuya Yamada$^{1,3}$\\
            {\tt ikuya@ousia.jp}
        \end{center}
    \end{minipage}
    \begin{minipage}{13em}
        \begin{center}
            Hiroyuki Shindo$^{2,3}$\\
            {\tt shindo@is.naist.jp}
        \end{center}
    \end{minipage}
    \\
    \\
    $^{1}$Studio Ousia, Tokyo, Japan\\
    $^{2}$Nara Institute of Science and Technology, Nara, Japan\\
    $^{3}$RIKEN AIP, Tokyo, Japan
}
\date{}

\begin{document}
    \maketitle
    \begin{abstract}
        This study proposes a \textit{Neural Attentive Bag-of-Entities} model, which is a neural network model that performs text classification using entities in a knowledge base.
        Entities provide unambiguous and relevant semantic signals that are beneficial for capturing semantics in texts.
        We combine simple high-recall entity detection based on a dictionary, to detect entities in a document, with a novel neural attention mechanism that enables the model to focus on a small number of unambiguous and relevant entities.
        We tested the effectiveness of our model using two standard text classification datasets (i.e., the 20 Newsgroups and R8 datasets) and a popular factoid question answering dataset based on a trivia quiz game.
        As a result, our model achieved state-of-the-art results on all datasets.
        The source code of the proposed model is available online at \url{https://github.com/wikipedia2vec/wikipedia2vec}.
    \end{abstract}

    \section{Introduction}

    Text classification is an important task, and its applications span a wide range of activities such as topic classification, spam detection, and sentiment classification.
    Recent studies showed that models based on neural networks can outperform conventional models (e.g., na\"{\i}ve Bayes) on text classification tasks \cite{kim:2014:EMNLP2014,iyyer-EtAl:2015,tang-qin-liu:2015:EMNLP,NIPS2015_5949,Jin2016,joulin-EtAl:2017:EACLshort,P18-1041}.
    Typical neural network-based text classification models are based on words.
    They typically use words in the target documents as inputs, map words into continuous vectors (embeddings), and capture the semantics in documents by using compositional functions over word embeddings such as averaging or summation of word embeddings, convolutional neural networks (CNN), and recurrent neural networks (RNN).

    Apart from the aforementioned approaches, past studies attempted to use entities in a knowledge base (KB) (e.g., Wikipedia) to capture the semantics in documents.
    These models typically represent a document by using a set of entities (or \textit{bag of entities}) relevant to the document \cite{Gabrilovich2006,Gabrilovich2007,Xiong2016}.
    The main benefit of using entities instead of words is that unlike words, entities provide unambiguous semantic signals because they are uniquely identified in a KB.
    One key issue here is to determine the way in which to associate a document with its relevant entities.
    An existing straightforward approach \cite{peng-liu-lin:2016:P16-1,Xiong2016} involves creating a set of relevant entities using an entity linking system to detect and disambiguate the names of entities in a document.
    However, this approach is problematic because
    (1) entity linking systems produce disambiguation errors \cite{Cornolti:2013:FBE:2488388.2488411}, and
    (2) entities appearing in a document are not necessarily relevant to the given document \cite{Gamon2013,dunietz-gillick:2014:EACL2014-SP}.

    This study proposes the \textit{Neural Attentive Bag-of-Entities} (NABoE) model, which is a neural network model that addresses the text classification problem by modeling the semantics in the target documents using entities in the KB.
    For each entity name in a document (e.g., \textit{``Apple''}), our model first detects entities that may be referred to by this name (e.g., \textit{Apple Inc.}, \textit{Apple (food)}), and then represents the document using the weighted average of the embeddings of these entities.
    The weights are computed using a novel neural attention mechanism that enables the model to focus on a small subset of the entities that are less ambiguous in meaning and more relevant to the document.
    In other words, the attention mechanism is designed to compute weights by jointly addressing entity linking and entity salience detection \cite{Gamon2013,dunietz-gillick:2014:EACL2014-SP} tasks.
    Furthermore, the attention mechanism improves the interpretability of the model because it enables us to inspect the small number of entities that strongly affect the classification decisions.

    We validate the effectiveness of our proposed model by addressing two important natural language tasks: a text classification task using two standard datasets (i.e., the 20 Newsgroups and R8 datasets), and a factoid question answering task based on a popular dataset derived from the \textit{quiz bowl} trivia quiz game.
    As a result, our model achieved state-of-the-art results on both tasks.
    The source code of the proposed model is available online at \url{https://github.com/wikipedia2vec/wikipedia2vec}.

    \section{Our Approach}

    Given a document, our model addresses the text classification task by using the following two steps: it first detects entities from the document, and then classifies the document using the proposed model with the detected entities as inputs.

    \subsection{Entity Detection}
    \label{subsec:entity-detection}

    In this step, we detect entities that may be relevant to the document.
    Here, we use a simple method based on an \textit{entity dictionary} that maps an entity name (e.g., ``Washington'') to a set of possible referent entities (e.g., \textit{Washington, D.C.} and \textit{George Washington}).
    In particular, we first take all words and phrases in a document, treat them as entity names if they exist in the dictionary, and detect all possible referent entities for each detected entity name.
    Following past work \cite{Hasibi2016,Xiong2016}, the boundary overlaps of the names are resolved by detecting only those that are the earliest and the longest.

    We use Wikipedia as the target KB, and the entity dictionary is built by using the names and their referent entities of all internal anchor links in Wikipedia \cite{StephenGuo}.
    We also collect two statistics from Wikipedia, namely \textit{link probability} and \textit{commonness} \cite{Mihalcea2007,Milne2008}. The former is the probability of a name being used as an anchor link in Wikipedia, whereas the latter is the probability of a name referring to an entity in Wikipedia.

    We generate a list of entities by concatenating all possible referent entities contained in the dictionary for each detected entity name, and feed it to the model presented in the next section.
    Note that we do not disambiguate entity names here, but detect all possible referent entities of the entity names.

    \subsection{Model}
    \label{subsec:model}

    \begin{figure}[t]
        \centering
        \includegraphics[width=220px,clip]{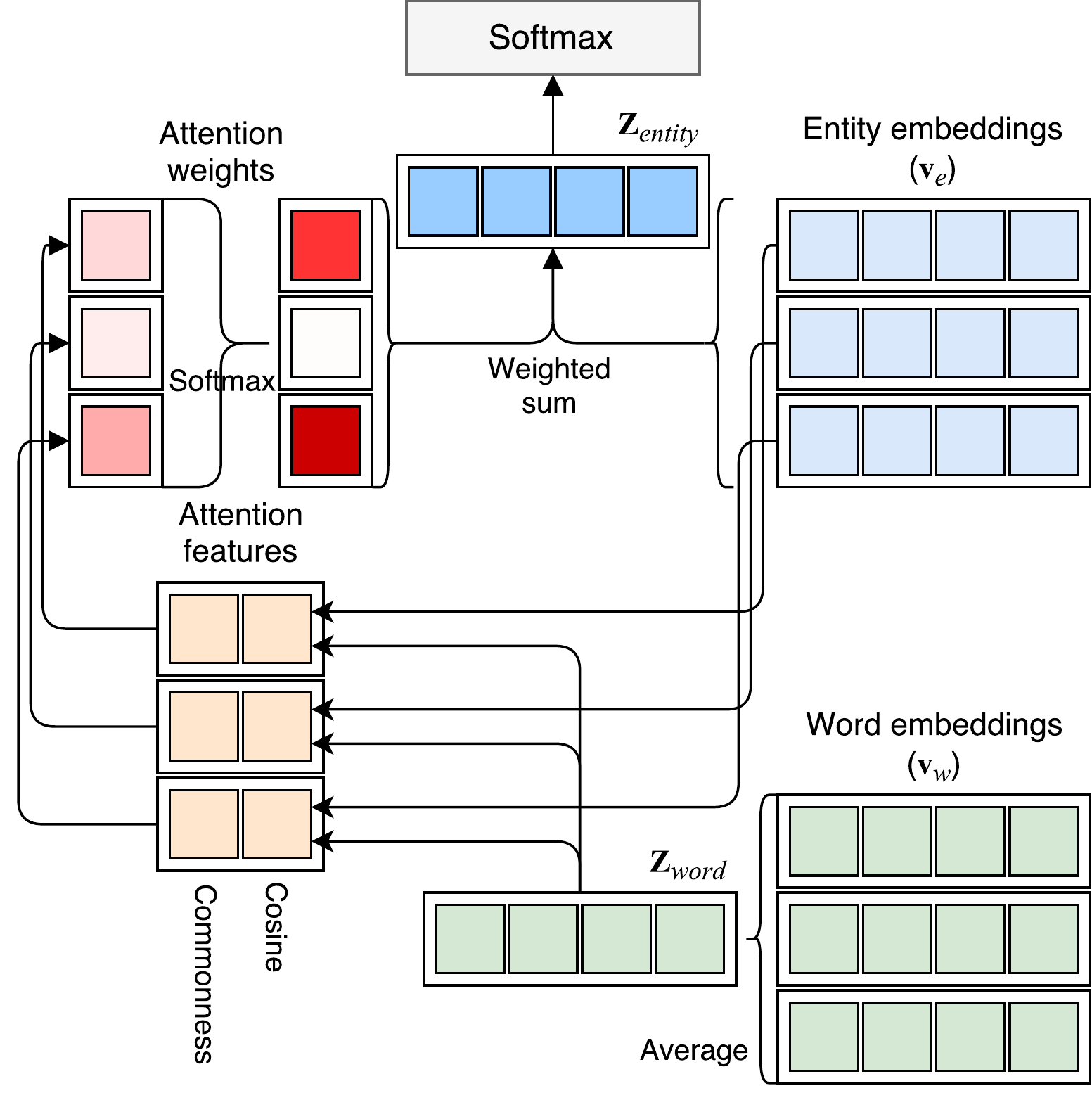}
        \caption{Architecture of the NABoE-entity model.}
        \label{fig:architecture}
    \end{figure}

    Figure \ref{fig:architecture} shows the architecture of our model.
    Given words $w_1, ..., w_N$, and entities $e_1, ..., e_K$ detected from target document $D$, we first compute the word-based representation of $D$:
    \begin{equation}
    \mathbf{z}_{word} = \frac{1}{N}\sum_{i=1}^{N} \mathbf{v}_{w_i},\;
    \end{equation}
    where $\mathbf{v}_{w} \in \mathbb{R}^d$ is the embedding of word $w$.
    We then derive the entity-based representation of $D$ as a weighted average of the embeddings of the entities:
    \begin{equation}
    \mathbf{z}_{entity} = \sum_{i=1}^{K} a_{e_i}\mathbf{v}_{e_i},\;
    \label{eq:entity-averaging}
    \end{equation}
    where $\mathbf{v}_{e} \in \mathbb{R}^d$ is the embedding of entity $e$ and $a_e$ the normalized attention weight corresponding to $e$ computed using the following softmax-based attention function:
    \begin{equation}
    a_e = \frac{\exp(\mathbf{w}_a^\top \Phi(e, D) + b_a)}{\sum_{i=1}^K\exp(\mathbf{w}_a^\top \Phi(e_i, D) + b_a)},
    \end{equation}
    where $\mathbf{w}_a \in \mathbb{R}^l$ is a weight vector, $b_a \in \mathbb{R}$ is the bias, and $\Phi(e, D)$ is a function that generates an $l$-dimensional vector consisting of the features of the attention function.

    We use the following two features in the attention function:
    \begin{itemize}
        \item \textbf{Cosine}: the cosine similarity between the embedding of the entity $\mathbf{v}_e$ and the word-based representation of the document $\mathbf{z}_{word}$.
        \item \textbf{Commonness}: the probability that the entity name refers to the entity in KB.
    \end{itemize}
    Here, our aim is to capture the relevance and the unambiguity of entity $e$ in document $D$ using the attention function.
    Thus, the problem is related to the tasks of entity salience detection \cite{Gamon2013,dunietz-gillick:2014:EACL2014-SP}, which aims to detect entities relevant (or salient) to the document, and entity linking, which aims to resolve the ambiguity of entities.
    The key assumption relating to these two tasks in the literature is that if an entity is semantically related to the given document, it is relevant to the document \cite{dunietz-gillick:2014:EACL2014-SP}, and it is likely to appear in the document \cite{Milne2008,Ratinov2011}.
    With this in mind and following past work \cite{Yamada2016}, we use the cosine similarity between $\mathbf{v}_e$ and $\mathbf{z}_{word}$ as a feature.
    Further, as in past entity linking studies, we also use the commonness of the name referring to the entity.

    Moreover, we derive a representation based both on entities and words by simply adding $\mathbf{z}_{entity}$ and $\mathbf{z}_{word}$\footnote{We also tested concatenating $\mathbf{z}_{entity}$ and $\mathbf{z}_{word}$ to derive $\mathbf{z}_{full}$; however, adding them generally achieved enhanced performance in our experiments presented below.}:
    \begin{equation}
    \mathbf{z}_{full} = \mathbf{z}_{entity} + \mathbf{z}_{word}.
    \label{eq:word-entity-representation}
    \end{equation}

    We then solve the task using a multiclass logistic regression classifier with the computed representation (i.e., with $\mathbf{z}_{entity}$ or $\mathbf{z}_{full}$) as features.
    In the remainder of this paper, we denote our models based on $\mathbf{z}_{entity}$ and $\mathbf{z}_{full}$ by \textbf{NABoE-entity} and \textbf{NABoE-full}, respectively.

    \section{Experimental Setup}

    In this section, we describe our experimental setup used both in the text classification and the factoid question answering experiments presented below.

    \subsection{Entity Detection}
    \label{subsec:kb-and-entity-dictionary}

    As the target KB, we used the September 2018 version of Wikipedia, which contains a total of 7,333,679 entities.\footnote{We downloaded the Wikipedia dump from Wikimedia Downloads: \url{https://dumps.wikimedia.org/}}
    Regarding the entity dictionary described in Section \ref{subsec:entity-detection}, we excluded an entity name if its link probability was lower than 1\% and a referent entity if its commonness given the entity name was lower than 3\% for computational efficiency.
    Entity names were treated as case-insensitive.
    As a result, the dictionary contained 18,785,550 entity names, and each name had 1.14 referent entities on average.

    Furthermore, to detect entities from a document, we also tested two publicly available entity linking systems, \textbf{Wikifier} \cite{Ratinov2011,cheng-roth:2013:EMNLP} and \textbf{TAGME} \cite{Ferragina2012}, instead of using dictionary-based entity detection.\footnote{In our experiments, we simply used all entities detected by the entity linking systems.}
    We selected these systems because they are capable of detecting non-named entities (e.g., technical terms) that are useful for addressing the text classification task.\footnote{In our preliminary experiments, we also tested three other state-of-the-art entity linking systems: AIDA \cite{Hoffart2011}, WAT \cite{Piccinno:2014:TWN:2633211.2634350}, and the commercial Entity Analysis API in Google's Cloud Language service.
        However, these systems achieved lower overall performance compared to Wikifier and TAGME because they tended to ignore non-named entities.
    }
    Here, we used the entities detected and disambiguated by these systems as inputs to our neural network model.

    \subsection{Pretrained Embeddings}

    We initialized the embeddings of words ($\mathbf{v}_w$) and entities ($\mathbf{v}_e$) using pretrained embeddings trained on KB.
    To learn embeddings from the KB, we used the method adopted in the open source Wikipedia2Vec tool \cite{Yamada2016,Yamada2018}.
    In particular, we generated an entity-annotated corpus from Wikipedia by treating entity links in Wikipedia articles as entity annotations, and trained skip-gram embeddings \cite{Mikolov2013,Mikolov2013a} of 300 dimensions with negative sampling using the generated corpus as inputs.
    The learned embeddings place similar words and entities close to one another in a unified vector space.
    Here, we used the same version of Wikipedia described in Section \ref{subsec:kb-and-entity-dictionary}.

    \section{Text Classification}
    \label{sec:text-classification}

    To evaluate the effectiveness of our proposed model, we first conducted the text classification task on two standard datasets, namely the 20 Newsgroups (20NG) \cite{Lang1995} and R8 datasets \cite{Debole2005}.

    \subsection{Setup}
    \label{subsec:text-classification-setup}

    Our experimental setup described in this section follows that in past work \cite{Liu:2015:TWE:2886521.2886657,Jin2016,C18-1016}.
    In particular, we used the 20NG and R8 datasets to train and test the proposed model.
    The 20NG dataset was created using the documents obtained from 20 Newsgroups and contained 11,314 training documents and 7,532 test documents.\footnote{We used the by-date version downloaded from the author's web site: \url{http://qwone.com/~jason/20Newsgroups/.}}
    The R8 dataset consisted of news documents from the eight most popular classes of the Reuters-21578 corpus \cite{Lewis1992} and comprised 5,485 training documents and 2,189 test documents.
    We created the development set for each dataset by selecting 5\% of the documents for training.
    Note that the class distribution of the R8 dataset is highly imbalanced.
    For example, the number of documents in the largest and smallest classes is 3,923 documents and 51 documents, respectively.

    We report the accuracy and macro-average F1 scores.
    The model was trained using mini-batch stochastic gradient descent (SGD) with its batch size set to 32 and its learning rate controlled by Adam \cite{kingma2014adam}.
    We used words and entities that were detected three times or more in the dataset and ignored the other words and entities.
    The size of the embeddings of words and entities was set to $d =300$.
    We used early stopping based on the accuracy of the development set of each dataset to avoid overfitting of the model.

    \subsection{Baselines}
    \label{subsec:text-classification-baselines}

    We used the following models as our baselines:

    \begin{itemize}
        \item \textbf{BoW-SVM} \cite{Jin2016}: This model is based on a conventional linear support vector machine (SVM) with bag of words (BoW) features.
        It outperformed the conventional na\"{\i}ve Bayes-based model.

        \item \textbf{BoE} \cite{Jin2016}:
        This model extends the skip-gram model;
        It learns different word embeddings per target class from the dataset, and a linear model based on learned word embeddings is used to classify the documents.
        The performance of this model was superior to that of many state-of-the-art models, including those based on the skip-gram and CBOW models \cite{Mikolov2013a}, and the paragraph vector model \cite{DBLP:conf/icml/LeM14}.

        \item \textbf{SWEM-concat} \cite{P18-1041}:
        This model is based on a neural network model with simple pooling operations (i.e., average and max pooling) over pretrained word embeddings.\footnote{We also tested all four models proposed in \newcite{P18-1041} (i.e., SWEM-aver, SWEM-max, SWEM-concat, and SWEM-hier). These models generally delivered comparable performance, with SWEM-concat slightly outperforming the other models on average.}
        Despite its simplicity, it outperformed many neural network-based models such as the word-based CNN model \cite{kim:2014:EMNLP2014} and RNN model with LSTM units \cite{P18-1041}.

        \item \textbf{TextEnt} \cite{C18-1016}:
        This model learns entity-aware document embeddings from Wikipedia, and uses a neural network model with the learned embeddings as pretrained parameters to address text classification.

    \end{itemize}

    As described in Section \ref{subsec:entity-detection}, we also tested the variants of our NABoE-entity and NABoE-full models for which \textbf{Wikifier} and \textbf{TAGME} were used as the entity detection methods.

    \subsection{Results}

    \begin{table}[tb]
        \centering
        \setlength\tabcolsep{3pt}
        \small{
        \begin{tabular}{l|cc|cc}
            & \multicolumn{2}{|c|}{20NG} & \multicolumn{2}{c}{R8}\\
            & Acc. & F1 & Acc. & F1\\
            \hline
            NABoE-entity          & .863          & .856          & .962          & .915 \\
            NABoE-entity w/o att. & .822          & .817          & .943          & .869 \\
            NABoE-entity w/o emb. & .844          & .838          & .957          & .892 \\
            NABoE-full            & \textbf{.868} & \textbf{.862} & \textbf{.971} & \textbf{.917} \\
            \hline
            Wikifier (NABoE-entity)          & .735 & .729 & .896 & .803\\
            Wikifier (NABoE-entity w/o att.) & .728 & .723 & .844 & .782\\
            Wikifier (NABoE-entity w/o emb.) & .727 & .722 & .861 & .755\\
            Wikifier (NABoE-full)            & .797 & .789 & .953 & .839\\
            \hline
            TAGME (NABoE-entity)          & .844 & .838 & .942 & .871\\ 
            TAGME (NABoE-entity w/o att.) & .826 & .821 & .924 & .857\\ 
            TAGME (NABoE-entity w/o emb.) & .842 & .836 & .942 & .865\\ 
            TAGME (NABoE-full)            & .860 & .853 & .958 & .889\\
            \hline
            BoW-SVM     & .790 & .783 & .947 & .851 \\
            BoE         & .831 & .827 & .965 & .886 \\
            SWEM-concat & .853 & .855 & .967 & .898 \\
            TextEnt     & .845 & .839 & .967 & .910 \\
        \end{tabular}
        }
        \caption{Results of the text classification task on the 20NG and R8 datasets.
            Here, \textit{w/o att.} and \textit{w/o emb.} represent the model without the neural attention mechanism and the model without the pretrained embeddings, respectively.}
        \label{tb:main-results}
    \end{table}

    Table \ref{tb:main-results} shows the results of our models and those of our baselines.
    Here, \textit{w/o att.} and \textit{w/o emb.} signify the model without the neural attention mechanism (all attention weights $a_e$ are set to $\frac{1}{K}$, where $K$ is the number of entities in the document) and the model without the pretrained embeddings (the embeddings are initialized randomly), respectively.

    Relative to the baselines, our models yielded enhanced overall performance on both datasets.
    The NABoE-full model outperformed all baseline models in terms of both measures on both datasets.
    Furthermore, the NABoE-entity model outperformed all the baseline models in terms of both measures on the 20NG dataset, and the F1 score on the R8 dataset.
    Moreover, our attention mechanism consistently improved the performance.
    These results clearly highlighted the effectiveness of our approach, which addresses text classification by using a small number of unambiguous and relevant entities detected by the proposed attention mechanism.
    Moreover, the pretrained embeddings improved the performance on both datasets.

    Further, the models based on the dictionary-based entity detection (see Section \ref{subsec:entity-detection}) generally outperformed the models based on the entity linking systems (i.e., Wikifier and TAGME).
    We consider that this is because these entity linking systems failed to detect or disambiguate entity names that were useful to address the text classification task.
    Moreover, our attention mechanism consistently improved the performance for Wikifier- and TAGME-based models because the attention mechanism enabled the model to focus on entities that were relevant to the document.

    \subsection{Analysis}

    \begin{table}[t]
        \centering
        \setlength\tabcolsep{1pt}
        \small{
        \begin{tabular}{l|cccc}
            Class & \begin{tabular}{c}\small SWEM\\-concat\end{tabular} & \begin{tabular}{c}\small NABoE\\-full\end{tabular} & \begin{tabular}{c}\small NABoE\\-entity\end{tabular} \\
            \hline
            \textbf{20NG:}\\
            \small{alt.atheism             } &         .780  & \textbf{.820} &         .804  \\
            \small{comp.graphics           } &         .787  &         .818  & \textbf{.822} \\
            \small{comp.os.ms-windows.misc } &         .746  &         .802  & \textbf{.811} \\
            \small{comp.sys.ibm.pc.hardware} &         .735  & \textbf{.754} &         .752  \\
            \small{comp.sys.mac.hardware   } &         .857  & \textbf{.865} &         .861  \\
            \small{comp.windows.x          } &         .837  &         .867  & \textbf{.870} \\
            \small{misc.forsale            } & \textbf{.854} &         .834  &         .805  \\
            \small{rec.autos               } &         .916  & \textbf{.929} &         .917  \\
            \small{rec.motorcycles         } &         .954  & \textbf{.968} &         .956  \\
            \small{rec.sport.baseball      } &         .946  & \textbf{.969} &         .966  \\
            \small{rec.sport.hockey        } &         .971  & \textbf{.981} &         .975  \\
            \small{sci.crypt               } & \textbf{.942} &         .940  &         .940  \\
            \small{sci.electronics         } &         .794  & \textbf{.806} &         .783  \\
            \small{sci.med                 } &         .878  &         .900  & \textbf{.905} \\
            \small{sci.space               } &         .921  & \textbf{.923} &         .918  \\
            \small{soc.religion.christian  } &         .905  & \textbf{.906} &         .905  \\
            \small{talk.politics.guns      } &         .826  & \textbf{.828} &         .819  \\
            \small{talk.politics.mideast   } &         .921  & \textbf{.940} &         .935  \\
            \small{talk.politics.misc      } &         .689  & \textbf{.694} &         .680  \\
            \small{talk.religion.misc      } &         .657  &         .702  & \textbf{.706}  \\

            \hline
            \textbf{R8:}\\
            \small{grain   } &         .750  & \textbf{.889} & \textbf{.889} \\
            \small{ship    } &         .781  &         .817  & \textbf{.822} \\
            \small{interest} &         .910  &         .885  &         .885  \\
            \small{money-fx} & \textbf{.909} &         .894  &         .898  \\
            \small{trade   } &         .894  & \textbf{.924} & \textbf{.924} \\
            \small{crude   } & \textbf{.971} &         .958  &         .954 \\
            \small{acq     } &         .979  & \textbf{.980} &         .966 \\
            \small{earn    } &         .989  & \textbf{.990} &         .980 \\
        \end{tabular}
        }
        \caption{Class-level F1 scores in each class on the 20NG and R8 datasets.}
        \label{tb:class-level-results}
    \end{table}

    \begin{table}[t]
        \centering
        \begin{tabular}{l|cc|cc}
            \multirow{2}{*}{} & \multicolumn{2}{c|}{20NG} & \multicolumn{2}{c}{R8}\\
            & Acc. & F1 & Acc. & F1\\
            \hline
            Commonness only           & .849          & .843          & .949          & .894 \\
            Cosine only               & .846          & .840          & .956          & .898 \\
            Both                      & \textbf{.863} & \textbf{.856} & \textbf{.962} & \textbf{.915} \\
        \end{tabular}
        \caption{Feature study of the neural attention mechanism of the NABoE-entity model.}
        \label{tb:attention-feature-study}
    \end{table}

    \begin{table*}[!htb]
        \centering
        \setlength\tabcolsep{3pt}
        \small{
        \begin{tabular}{p{3.5cm}|p{12.1cm}}
            Class & Top entities \\
            \hline
            \small{\textbf{20NG:}}\\
            \small{alt.atheism               } & \small{Christian ethics, Atheism, Moral agency, Gregg Jaeger, Fred Rice}\\
            \small{comp.graphics             } & \small{Algorithm, Ray tracing (graphics), Framebuffer, Image file formats, TIFF}\\
            \small{comp.os.ms-windows.misc   } & \small{Windows 3.1x, Microsoft Windows, Windows NT, CONFIG.SYS, BMP file format}\\
            \small{comp.sys.ibm.pc.hardware  } & \small{BIOS, Don't Copy That Floppy, SCSI host adapter, Nonvolatile BIOS memory, Parallel SCSI}\\
            \small{comp.sys.mac.hardware     } & \small{PowerBook, Macintosh Quadra 610, Macintosh Quadra 650, FirstClass, Macintosh SE/30}\\
            \small{comp.windows.x            } & \small{X-Perts, Xterm, OPEN LOOK, OpenWindows, Man page}\\
            \small{misc.forsale              } & \small{Freight transport, Make Me an Offer, AC adapter, Plaque reduction neutralization test, Outline of working time and conditions}\\
            \small{rec.autos                 } & \small{Manual Shift, Chassis, Automotive industry, Nissan, Ford Probe}\\
            \small{rec.motorcycles           } & \small{United States Department of Defense, Motorcycle, ZX8302, Honda motorcycles, Pillion, Hawk GT}\\
            \small{rec.sport.baseball        } & \small{Pitcher, Inning, The Jays, Home run, Bullpen}\\
            \small{rec.sport.hockey          } & \small{National Hockey League, Goaltender, ESPN, The Penguins, Achkar}\\
            \small{sci.crypt                 } & \small{Cryptography, Algorithm, Escrow, Considered harmful, Encryption}\\
            \small{sci.electronics           } & \small{Solvent, Copy protection, Electronics, Lead–acid battery, Printed circuit board}\\
            \small{sci.med                   } & \small{Infection, Antibiotics, Kirlian photography, Allergy, Kirlian}\\
            \small{sci.space                 } & \small{Spacecraft, SunOS, Vandalism, VIA International, Space station}\\
            \small{soc.religion.christian    } & \small{Rutgers University, Geneva, Byler, Immaculate Conception, Original sin}\\
            \small{talk.politics.guns        } & \small{Ranch, BD's Mongolian Grill, Firearm, Second Amendment to the United States Constitution, Feustel}\\
            \small{talk.politics.mideast     } & \small{Serdar Argic, Israelis, Palestinians, Palestine Liberation Organization, Arabs}\\
            \small{talk.politics.misc        } & \small{Clayton Cramer, Janet Reno, Police state, Ronzone, Federal Bureau of Investigation}\\
            \small{talk.religion.misc        } & \small{Christian ethics, Thomas George Lanphier, David Koresh, Albert Sabin, Josephus}\\

            \hline
            \small{\textbf{R8:}}\\
            \small{grain   } & \small{Grain, Tonne, Price support, Oil reserves, United States Senate}\\
            \small{ship    } & \small{Freight transport, Shipbuilding, Flag of convenience, Cargo, Persian Gulf}\\
            \small{trade   } & \small{Balance of trade, Export, International trade, Economic sanctions, Import}\\
            \small{interest} & \small{Interest rate, Prime rate, Repurchase agreement, Balance of trade, Money market}\\
            \small{money-fx} & \small{Exchange rate, Currency, Money market, Foreign exchange market, Monetary policy}\\
            \small{crude   } & \small{Petroleum, West Texas Intermediate, Price of oil, OPEC, Oil platform}\\
            \small{acq     } & \small{Common stock, Tender offer, Privately held company, Preferred stock, Shares outstanding}\\
            \small{earn    } & \small{QTR, Dividend, Stock split, Net profit, Income fund}\\
        \end{tabular}
        }
        \caption{Top five influential entities for each class of the NABoE-entity model in the 20NG and R8 datasets.}
        \label{tb:top-entities}
    \end{table*}

    In this section, we provide a detailed analysis of the performance of our model in terms of conducting the text classification task.
    We first provide a comparison of the SWEM-concat, NABoE-entity, and NABoE-full models using class-level F1 scores on both of the datasets (see Table \ref{tb:class-level-results}).
    Here, we aim to compare the detailed performance of the word-based model (SWEM-concat), entity-based model (NABoE-entity), and the model based on both words and entities (NABoE-full).
    Compared with the SWEM-concat model, the NABoE-full and NABoE-entity models performed more accurately in 23 out of 28 and 17 out of 28 classes, respectively.
    This result clearly demonstrates the ability of the model to successfully capture strong semantic signals that can only be obtained from entities.
    Moreover, we observed that the NABoE-entity model achieved weaker performance especially for the \textit{misc.forsale} class in the 20NG dataset and several classes in the R8 dataset.
    Regarding the \textit{misc.forsale} class, because documents in this class contain a wider variety of entities (i.e., objects users want to sell) than other classes, the model failed to capture the effective semantic signals from the entities.
    Further, as described in the error analysis provided below, it often appeared to be difficult to distinguish pairs of similar classes in the R8 dataset based only on entities.

    Next, we conducted a feature study of the attention mechanism by excluding one feature at a time from the NABoE-entity model (Table \ref{tb:attention-feature-study}).
    We found both of the features to make an important contribution to the performance.

    Furthermore, to investigate the attention mechanism in more detail, we computed the top influential entities in the attention mechanism for each class on the 20NG and R8 datasets.
    In particular, we calculated the number of times each entity obtained the highest attention weight in the test documents in each class and selected the five most frequent ones.
    Table \ref{tb:top-entities} presents the results.
    Overall, our attention mechanism successfully selected entities that were highly relevant to each class.
    For example, \textit{Cryptography}, \textit{Algorithm}, \textit{Escrow}, \textit{Considered harmful}, and \textit{Encryption} were selected for the \textit{sci.crypt} class.
    Furthermore, although we did not explicitly perform entity disambiguation, the model successfully overcame the ambiguity issues in the entity names and attended to the entities that were relevant to the classes.

    Subsequently, we conducted an error analysis by selecting 50 random test documents for which the NABoE-entity model made wrong predictions.
    Most of the errors were caused by two pairs of classes:
    22 errors were caused by misclassifying documents of \textit{acq} (corporate acquisitions) and those of \textit{earn} (corporate earnings), and 13 errors were caused by misclassifying documents of \textit{interest} and those of \textit{money-fx}.
    Furthermore, the model tended to perform poorly if a document contained entities that strongly indicate an incorrect class.
    For example, a \textit{money-fx} document containing the entity \textit{interest rate} multiple times was classified into the \textit{interest} class, and a document in the \textit{acq} class reporting news related to oil companies (i.e., ExxonMobil and ZENEX) was classified into the \textit{crude} class.

    \section{Factoid Question Answering}

    In this section, we address factoid question answering based on a dataset consisting of questions of the \textit{quiz bowl} trivia quiz game.
    Factoid question answering is one of the common settings of question answering that aims to predict an entity (e.g., events, authors, and books) that is described in a given question.
    The players of quiz bowl solve questions consisting of sentences that describe an entity.
    Quiz bowl questions have frequently been used for evaluating neural network-based models in recent studies \cite{iyyer-EtAl:2014:EMNLP2014,iyyer-EtAl:2015,TACL1065}.

    This task has a significantly larger number of target classes compared to the task addressed in the previous experiment.
    Our main aim here is to evaluate the effectiveness of using entities to capture the finer-grained semantics required to perform the task of factoid question answering effectively.

    \subsection{Setup}

    Our experimental setup described in this section follows that in past work \cite{Xu2016,TACL1065}.
    We address this task as a text classification problem that selects the most relevant answer from the possible answers observed in the dataset.
    We obtained the dataset proposed in \newcite{iyyer-EtAl:2014:EMNLP2014}\footnote{This dataset was downloaded from the authors' web page: \url{https://cs.umd.edu/˜miyyer/qblearn/}.}.
    We only used questions in the history and literature categories.
    Furthermore, we excluded questions of which the answers appear fewer than six times in the dataset.
    As a result, the number of candidate answers was 303 and 424 in the history and literature categories, respectively.
    We used 20\% of questions each for the development set and test sets, and the remaining 60\% for the training set.
    As a result, the training, development, and test sets consisted of 1,535, 511, and 511 questions for the history category, and 2,524, 840, and 840 questions for the literature category.

    The settings we used to train the model were the same as those in the previous experiment (see Section \ref{subsec:text-classification-setup}).
    The model was trained using mini-batch SGD with its learning rate controlled by Adam \cite{kingma2014adam} and its mini-batch size set to 32.
    We used words and entities that were detected three times or more in the dataset, and ignored the other words and entities.
    The size of the embeddings of words and entities was set to $d =300$.
    As in past work, we report the accuracy score, and the score on the development set was used for early stopping.

    \subsection{Baselines}
    We used the following baseline models:

    \begin{itemize}
        \item \textbf{BoW} \cite{Xu2016} This model is based on a logistic regression classifier with conventional binary BoW features.
        \item \textbf{FTS-BRNN} \cite{Xu2016} This model is based on a bidirectional RNN with gated recurrent units (GRU). It uses the logistic regression classifier with the features derived by the RNN.
        \item \textbf{NTEE} \cite{TACL1065} This model is a state-of-the-art model that uses a multi-layer perceptron classifier with the features computed using the embeddings of words and entities trained on Wikipedia using the neural network model proposed in their paper.
    \end{itemize}

    Similar to our previous experiment, we also add \textbf{SWEM-concat}, and the variants of our NABoE-entity and NABoE-full models based on \textbf{Wikifier} and \textbf{TAGME} (see Section \ref{subsec:text-classification-baselines}).
    Note that all the baselines address the task as a text classification problem.

    \subsection{Results and Analysis}

    \begin{table}[tb]
        \centering
        \setlength\tabcolsep{3.5pt}
        \small{
        \begin{tabular}{l|cc}
            Name & History & Literature\\
            \hline
            NABoE-full                & \textbf{.949} & \textbf{.985}\\
            NABoE-entity              &         .941  &         .979 \\
            NABoE-entity w/o att.     &         .845  &         .943 \\
            NABoE-entity w/o emb.     &         .941  &         .973 \\
            \hline
            Wikifier (NABoE-full)            &         .935  &         .967\\
            Wikifier (NABoE-entity)          &         .930  &         .952\\
            Wikifier (NABoE-entity w/o att.) &         .924  &         .941\\
            Wikifier (NABoE-entity w/o emb.) &         .934  &         .949\\
            \hline
            TAGME (NABoE-full)               &         .941  &         .977\\
            TAGME (NABoE-entity)             &         .930  &         .963\\
            TAGME (NABoE-entity w/o att.)    &         .922  &         .961\\
            TAGME (NABoE-entity w/o emb.)    &         .932  &         .962\\
            \hline
            BoW                        &         .508  &         .462\\
            FTS-BRNN                   &         .881  &         .931\\
            NTEE                       &         .947  &         .951\\
            SWEM-concat                &         .900  &         .966\\
        \end{tabular}
        }
        \caption{Accuracy of the proposed and baseline methods for the factoid QA task.}
        \label{tb:qb-results}
    \end{table}

    Table \ref{tb:qb-results} provides the results of our models and those of our baselines.
    Overall, our models achieved enhanced performance on this task.
    In particular, the NABoE-full model successfully outperformed all the baseline models, and the NABoE-entity model achieved competitive performance and outperformed all the baseline models in the literature category.
    These results clearly highlighted the effectiveness of our model for this task.

    Furthermore, similar to the previous text classification experiment, the attention mechanism and the pretrained embeddings consistently improved the performance.
    Moreover, the models based on dictionary-based entity detection outperformed the models based on the entity linking systems.

    We also conducted an error analysis using the NABoE-entity model and the test questions in the history category.
    We found nearly 70\% of the errors to be caused by questions of which the answers were country names.
    This is because these questions tended to provide indirect clues (e.g., describing a notable person born in the country) and most entities used in these clues do not directly indicate the answer (i.e., country names).
    Furthermore, our model failed in difficult cases such as predicting \textit{Tokugawa shogunate} instead of \textit{Tokugawa Ieyasu}.

    \section{Related Work}

    KB entities have been conventionally used to model the semantics in texts.
    A representative example is Explicit Semantic Analysis (ESA) \cite{Gabrilovich2006,Gabrilovich2007}, which represents a document using a bag of entities, namely a sparse vector of which each dimension corresponds to the relevance score of the text to each entity.
    This simple method is shown to be effective for various NLP tasks including text classification \cite{Gabrilovich2006,Gupta:2008:TCK:1620163.1620203,negi-rosner:2013:SemEval-2013} and information retrieval \cite{Egozi:2011:CIR:1961209.1961211,Xiong2016},

    Several neural network models that use KB entities to capture the semantics in texts have been proposed.
    These models typically depend on an additional preprocessing step that extracts the relevant entities from the target texts.
    For example, \newcite{ijcai2017-406} used the Probase conceptualization API for short text classification by retrieving the Probase entities that were relevant to the target text and used them in a model based on CNN.
    \newcite{pilehvar-EtAl:2017:Long} also extracted entities using a graph-based linking algorithm and used these entities in a neural network model.
    A similar approach was adopted in \newcite{C18-1016,10.1007/978-3-319-94042-7_10}; they extracted entities from the target text using an entity linking system and simply used the detected entities in a neural network model.
    However, unlike these models, our proposed model addresses the task in an \textit{end-to-end} manner; i.e.,
    entities that are relevant to the target text are automatically selected using our neural attention mechanism.
    Furthermore, we also used the model proposed by \newcite{C18-1016} as a baseline in our text classification experiments.

    Additionally, our work is also related to studies on entity linking.
    Entity linking models can be roughly classified into two groups: \textit{local} models, which resolve entity names independently using the contextual relevance of the entity given a document, and \textit{global} models, in which all the entity names in a document are resolved simultaneously to select a topically coherent set of results \cite{Ratinov2011}.
    Recent state-of-the-art models typically combine both of these models \cite{Yamada2016,ganea-hofmann:2017:EMNLP2017,cao-EtAl:2018:C18-1,kolitsas-ganea-hofmann:2018:K18-1}.
    However, several studies also showed that the local model alone can achieve results competitive to those of the global and combined models \cite{eshel-EtAl:2017:CoNLL,ganea-hofmann:2017:EMNLP2017,TACL1065,cao-EtAl:2018:C18-1,kolitsas-ganea-hofmann:2018:K18-1}.
    In this study, we adopt a simple but effective local model, which uses cosine similarity between the embedding of the target entity and the word-based representation of the document to capture the relevance of an entity given a document.

    \section{Conclusions}
    This study proposed NABoE, which is a neural network model that performs text classification using entities in Wikipedia.
    We combined simple dictionary-based entity detection with a neural attention mechanism to enable the model to focus on a small number of unambiguous and relevant entities in a document.
    We achieved state-of-the-art results on two important NLP tasks, namely text classification and factoid question answering, which clearly verified the effectiveness of our approach.
    As a future task, we intend to more extensively analyze our model and explore its effectiveness for other NLP tasks.
    Furthermore, we would also like to test more expressive neural network models for example by integrating global entity coherence information into our neural attention mechanism.

    \bibliographystyle{acl_natbib}
    \bibliography{library}
\end{document}